\documentclass[10pt,twocolumn,letterpaper]{article}

\usepackage{iccv}              

\definecolor{iccvblue}{rgb}{0.21,0.49,0.74}
\usepackage[pagebackref,breaklinks,colorlinks,allcolors=iccvblue]{hyperref}
\usepackage{array}
\usepackage{multirow}

\def\paperID{*****} 
\def\confName{ICCV}
\def\confYear{2025}

\title{SemiSegECG: A Multi-Dataset Benchmark for Semi-Supervised Semantic Segmentation in ECG Delineation}

\author{
Minje Park$^*$, Jeonghwa Lim$^*$, Taehyung Yu, and Sunghoon Joo$^\dag$ \\
VUNO Inc.\\
{\tt\small \{minje.park, jeonghwa.lim, taehyung.yu, sunghoon.joo\}@vuno.co}
}

\begin{document}
\maketitle
\begin{abstract}
Electrocardiogram (ECG) delineation, the segmentation of meaningful waveform features, is critical for clinical diagnosis. Despite recent advances using deep learning, progress has been limited by the scarcity of publicly available annotated datasets. Semi-supervised learning presents a promising solution by leveraging abundant unlabeled ECG data. In this study, we present \textbf{\textsf{SemiSegECG}}, the first systematic benchmark for semi-supervised semantic segmentation (SemiSeg) in ECG delineation. We curated and unified multiple public datasets, including previously underused sources, to support robust and diverse evaluation. We adopted five representative SemiSeg algorithms from computer vision, implemented them on two different architectures: the convolutional network and the transformer, and evaluated them in two different settings: in-domain and cross-domain. Additionally, we propose ECG-specific training configurations and augmentation strategies and introduce a standardized evaluation framework. Our results show that the transformer outperforms the convolutional network in semi-supervised ECG delineation. We anticipate that \textbf{\textsf{SemiSegECG}} will serve as a foundation for advancing semi-supervised ECG delineation methods and will facilitate further research in this domain.
\end{abstract}

{\let\thefootnote\relax\footnotetext{$*$ Equal contribution.}\par}
{\let\thefootnote\relax\footnotetext{$\dag$ Corresponding author.}\par}
\section{Introduction}
\label{sec:intro}

The electrocardiogram (ECG) is a non-invasive, widely used tool for monitoring cardiac electrical activity. A typical ECG waveform includes the P wave, QRS complex, and T wave, representing atrial depolarization, ventricular depolarization, and ventricular repolarization, respectively. Accurate segmentation of these components—termed ECG delineation—is critical for clinical diagnosis~\citep{gacek2011ecg}.

\begin{figure}[t]
  \centering
  \includegraphics[width=3.28125in]{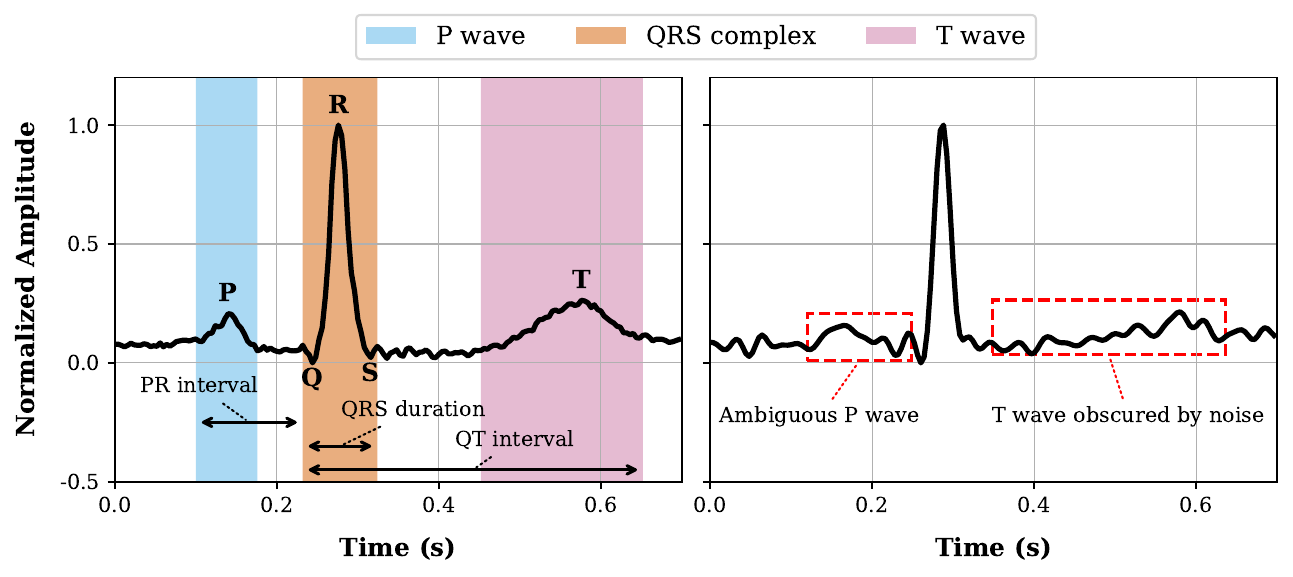}
  \caption{Examples of clean (left panel) and noisy (right panel) ECG recordings. In the left panel, standard waveform components (P wave, QRS complex, and T wave) are annotated, together with the PR interval, QRS duration, and QT interval. In the right panel, an ambiguous P wave and a noise-masked T wave make their boundaries difficult to identify.}
  \label{fig:deli-clean_noisy}
  \vspace{-0.5cm}
\end{figure}

Traditional methods using signal processing (e.g., wavelet transforms~\citep{martinez2004wavelet}) often struggle with signal variability and noise~\citep{elgendi2014revisiting, liu2018performance}, as shown in Figure~\ref{fig:deli-clean_noisy}. Deep learning has recently shown promise, treating delineation as a semantic segmentation task~\citep{jimenez2019u, liang2022ecg_segnet, chen2023post, chen2024ecgvednet, joung2024deep, jimenez2024generalising}, but existing models rely on small-scale datasets due to costly expert labeling.

Semi-supervised semantic segmentation (SemiSeg) can bridge this gap by utilizing abundant unlabeled ECG data~\citep{goldberger2000physiobank}. In computer vision, SemiSeg methods such as consistency regularization and self-training have proven effective~\citep{pelaez2023survey}. Nevertheless, their application to ECG delineation faces two critical challenges: (1) the absence of standardized benchmarks, and (2) insufficient evaluation in realistic ECG scenarios.

To address these challenges, we:
\begin{itemize}
    \item Introduce \textbf{\textsf{SemiSegECG}}, the first standardized benchmark for semi-supervised ECG delineation.
    \item Curate and integrate multiple ECG datasets, including previously underutilized public resources, and develop ECG-specific augmentation and training strategies.
    \item Evaluate five representative SemiSeg algorithms across varying label availability and distribution shift, considering both the segmentation accuracy and clinically relevant interval error metrics.
\end{itemize}

\begin{table*}[t]
    \centering
    \scriptsize
    \caption{
    ECG database characteristics. Lead types are defined as: 12-lead$=$standard limb (I, II, III, aVR, aVL, aVF) + precordial (V1--V6) leads; 6-lead$=$limb-only; 2-lead$=$two selected leads (e.g., MLII). Label types are lead-specific (separate onset-offset per lead), integrated (single annotation across all leads), and interval-only (PR, QRS, QT intervals without delineation annotation).
    }
    \vspace{-0.1cm}
    \begin{tabular}{l|c|c|c|c|c|c|c|ccc}
    \toprule
    \multirow{2}{*}{\textbf{Source}}&  \multirow{2}{*}{\textbf{\# Subjects}}&\multirow{2}{*}{\textbf{\# ECGs}} & \multirow{2}{*}{\textbf{Duration (labeled)}}& \multirow{2}{*}{\textbf{Sample rate}}& \multirow{2}{*}{\textbf{Lead type}}&\multirow{2}{*}{\textbf{Label type}}& \multicolumn{4}{c}{\textbf{\# Samples}}\\
    &   && & &  &&  All&Train& Validation&Test\\
    \midrule
    \textit{LUDB} &  200&200  & 10 s& 500 Hz & 12-lead   & Lead-specific &  2,369&1,427& 468&474\\
    \textit{QTDB} &  105&105  & 5.9--253.6 s& 250 Hz & 2-lead   & Integrated &  718&422& 148&148\\
    \textit{ISP} &  499&499  & 10 s& 1000 Hz & 12-lead   & Integrated &  5,988&3,792& 1,272&924\\
    \textit{Zhejiang} &  334&334  & 1.3--7.1 s& 2000 Hz & 12-lead   & Integrated &  4,008&2,400& 804&804\\
    \textit{PTB-XL} &  18,885&21,837  & 10 s& 500 Hz &12-lead   & -&  270,085&270,085& -&-\\
    \textit{mECGDB} &  205&205  & 2--10 s& 250 Hz & 6-lead & Interval only &  205&-& -&205\\
    \bottomrule
    \end{tabular}
    \label{tab:ecg_dataset}
\end{table*}

\section{Benchmark Design}
\label{sec:bench}

\subsection{ECG Databases}
Five public ECG databases were curated for benchmarking, with one additional private database. The characteristics of the databases are summarized in Table~\ref{tab:ecg_dataset}. \textit{LUDB}~\citep{kalyakulina2020ludb} and \textit{QTDB}~\citep{laguna1997database} have served as standard resources in previous ECG delineation studies~\citep{jimenez2021delineation,liang2022ecg_segnet,chen2023post}. In contrast, \textit{ISP}~\citep{avetisyan2024ispdataset} and \textit{Zhejiang}~\citep{zheng202012} are relatively new and remain underutilized in delineation research. These four databases, which provide ground-truth delineation annotations, served as the core training resources. Each database (\textit{LUDB}, \textit{QTDB}, \textit{ISP}, \textit{Zhejiang}) was split subject-wise randomly into training, validation, and test sets (6:2:2 ratio except for \textit{ISP} which provides an official split). \textit{PTB-XL}~\citep{wagner2020ptb}, a large-scale benchmark database for ECG classification, was used as an out-domain unlabeled dataset. We also utilized a private mobile ECG database (\textit{mECGDB}), which consists of 6-lead ECGs measured by portable devices in a non-clinical setting unlike the others, to evaluate model generalization under distribution shift.

\subsection{Data Preprocessing}
Each ECG lead was treated as an independent training instance, considering its unique spatial orientation and waveform morphology~\citep{gacek2011ecg}, while also enlarging the training set. Labeled ECGs were cropped or zero-padded to a fixed 10-second length, reflecting the standard recording window for resting ECGs in routine clinical practice~\citep{van2018normal,seo2022multiple}. All signals were resampled to 250 Hz, which is the lowest native rate among the benchmark databases, to ensure uniform temporal resolution and prevent artifacts introduced by upsampling. A bandpass filter (0.67--40 Hz) was applied to remove baseline drift and high-frequency noise. Z-score normalization was applied to signals before model input to improve training stability.

Official delineation labels were used for \textit{LUDB} and \textit{ISP}. For \textit{QTDB} and \textit{Zhejiang}, we used the label set released by a previous study~\citep{jimenez2024generalising}. This label set compensates for missing beats and incomplete onset-offset annotations in \textit{QTDB} and includes new delineation labels for \textit{Zhejiang}. \textit{mECGDB} contains only clinically relevant interval labels, without explicit delineation labels. All the labels were annotated by at least one expert cardiologist.

\subsection{Benchmarking Protocol}
\textbf{\textsf{SemiSegECG}} evaluates algorithm performance under two distinct conditions: in-domain and cross-domain settings.

\textbf{In-domain setting}.
 The in-domain setting reflects a typical use case where both labeled and unlabeled data come from the same source. We simulated semi-supervised conditions by randomly selecting labeled training subsets (1/16, 1/8, 1/4, 1/2) and using the entire training set as unlabeled data. Models were independently trained and evaluated per dataset and label proportion.

\textbf{Cross-domain setting}.
 The cross-domain setting reflects a practical scenario involving heterogeneous sources across labeled, unlabeled, and potentially test data. The four labeled databases were merged into a unified dataset, preserving the original splits.
\textit{PTB-XL} served as an out-domain unlabeled dataset. Models were evaluated on both the merged in-domain test set and \textit{mECGDB}, testing generalization across distribution shifts originating from different measuring environments (e.g., device types).

We compared convolutional~\citep{krizhevsky2012imagenet} and Transformer-based~\citep{vaswani2017attention} encoders, ResNet~\citep{he2016deep} and Vision Transformer (ViT)~\citep{dosovitskiy2021an}, selected for their proven reliability in image and ECG tasks~\citep{chen2017deeplab, zheng2021rethinking, ribeiro2020automatic, na2024guiding}. They were paired with a lightweight fully convolutional network (FCN) decoder~\citep{long2015fully}. Performance metrics included mean intersection over union (mIoU) for segmentation accuracy and mean absolute error (MAE) of ECG intervals (PR, QRS, QT) for clinical validity. Results were obtained from the model checkpoints that achieved the highest mIoU on the validation set.

\section{Experiments}

\subsection{SemiSeg Algorithms}
We benchmarked five SemiSeg algorithms originally developed for computer vision, each representing a distinct learning paradigm: \textbf{Mean Teacher (MT)}~\citep{tarvainen2017mean}, \textbf{FixMatch}~\citep{sohn2020fixmatch}, \textbf{Cross Pseudo Supervision (CPS)}~\citep{chen2021semi}, \textbf{Regional Contrast (ReCo)}~\citep{liu2022bootstrapping}, and \textbf{Self-Training++ (ST++)}~\citep{yang2022st++}.

\textbf{MT}. A student model learns to match predictions from a teacher model whose weights are updated as an exponential moving average (EMA) of the student’s, providing stable pseudo-labels to guide the student's predictions.

\textbf{FixMatch}. High-confidence predictions from weak augmentations of unlabeled data are used as pseudo-labels to supervise the same inputs with strong augmentations. This encourages robustness across perturbations.

\textbf{CPS}. Two models generate pseudo-labels for each other and are trained mutually, promoting consistency and reducing overfitting through regularized disagreement.

\textbf{ReCo}. Region-level contrastive regularization is applied to pixel embeddings obtained via an additional projection head from the encoder. Pixels with confidence scores between the easy and hard thresholds are selected as queries. These queries are pulled toward class prototypes of their predicted label and pushed away from those of others, sharpening segmentation boundaries.

\textbf{ST++}. With multi-step training, pseudo-labeled data is gradually introduced into training based on confidence and a predefined schedule. Early training focuses on labeled data, reducing noise from uncertain pseudo-labels.

A supervised baseline (\textbf{Scratch}) trained only with labeled data was included for comparison.

\subsection{Augmentation Strategy Exploration}
Effective augmentation is crucial for semi-supervised learning, enabling better utilization of labeled and unlabeled data~\citep{zhao2023augmentation}. However, conventional image-based strategies may not be well-suited for ECG signals~\citep{ rahman2023systematic, lim2025specialized}, distorting ECG-specific characteristics. To address this, we explored augmentation policies tailored to ECG segmentation. Augmentations were grouped as weak (minor global changes used to create pseudo-labels) or strong (larger perturbations applied to training inputs, yet preserving signal structure). Guided by prior studies~\citep{nonaka2020data, lee2022efficient}, we chose:
\begin{itemize}
    \item \textbf{Weak}: random resized cropping and horizontal flipping.
    \item \textbf{Strong}: baseline shift, powerline noise, amplitude scaling, sine-wave noise, and white noise.
\end{itemize}

Optimal weak and strong augmentation strategies were identified on \textit{LUDB} at the 1/16 label ratio with a ResNet backbone (\textbf{Scratch} for weak, \textbf{FixMatch} for strong) and then applied to all the algorithms.

\subsection{Implementation Details}
To ensure a fair benchmark, the following configurations were applied commonly across the experiments. The details are available in the public code repository\footnote{https://github.com/bakqui/semi-seg-ecg}.

\textbf{Model architecture}
We adopted compact encoders: ResNet-18~\citep{he2016deep} and ViT-Tiny~\citep{touvron2021training}, chosen to prevent overfitting on our modest-sized datasets while keeping parameter counts comparable. The decoder was implemented as a two-layer FCN consisting of a hidden dimension of 128 and a dropout layer~\citep{srivastava2014dropout} with $p=0.1$.

\textbf{Training schedule}.
All models were trained for 100 epochs with a batch size of 16. We used the AdamW optimizer~\citep{loshchilov2018decoupled} with a weight decay of 0.05. The learning rate followed a cosine annealing schedule~\citep{loshchilov2017sgdr}, warming up from 0 to 0.001 over the first 10 epochs and gradually decreasing to 0.0001.

\textbf{Augmentation}.
All augmentations were applied with a selection probability of 0.5, except for random resized cropping. For strong augmentation, we adopted the RandAugment policy~\citep{cubuk2020randaugment} with the searched pool of the augmentations.

\textbf{SemiSeg-specific hyperparameters}.
 Hyperparameter tuning was conducted on \textit{LUDB} at the 1/16 label ratio using a ResNet-18. For EMA-based models, the decay rate was set to 0.99. The confidence threshold was 0.8 for \textbf{FixMatch}, and 0.65 (easy) and 0.8 (hard) for \textbf{ReCo}. The projection head used in \textbf{ReCo} consisted of two convolutional layers with 128 channels each.
\begin{figure}[h]
  \centering
  \includegraphics[width=3.28125in]{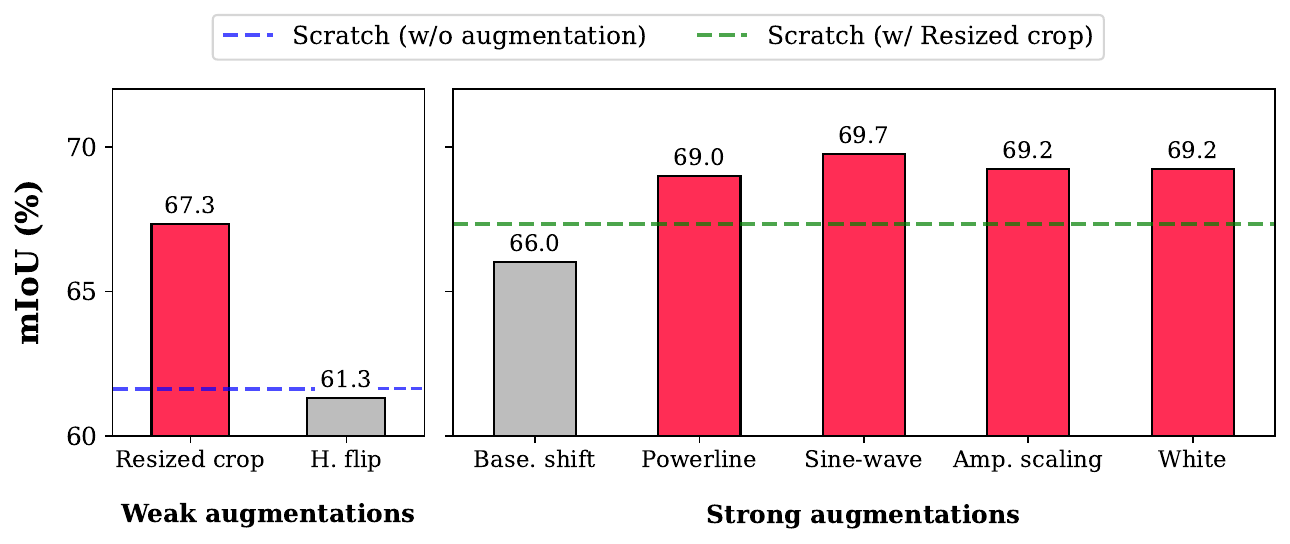}
  \caption{
  Comparison of augmentation strategies. Left panel: mIoU of supervised models (Scratch) using weak augmentations. Right panel: FixMatch models with fixed random resized cropping and one of the strong augmentations. Dashed lines indicate Scratch performance without any augmentation (blue) and with random resized cropping (green).
  }
  \label{fig:aug_ablation}
  \vspace{-0.5cm}
\end{figure}

\section{Results}

\subsection{Optimal Augmentation Strategy}
We empirically determined the optimal augmentation strategies for all subsequent experiments. Figure~\ref{fig:aug_ablation} illustrates their impact on delineation performance. For weak augmentation, random resized cropping significantly improved \textbf{Scratch} performance, while horizontal flipping degraded performance, likely because reversing the signal order (P-QRS-T becomes T-QRS-P) confused the model’s temporal cues. For strong augmentation, baseline shift reduced the performance of \textbf{FixMatch}, but powerline noise, sine-wave noise, amplitude scaling, and white noise all led to improvements. The best results arose when RandAugment combined any three of these four strong augmentations per sample ($\text{mIoU}=70.9\%$). Accordingly, we adopted random resized cropping as the weak augmentation and the four-operation RandAugment policy as the strong configuration for all later experiments.

\begin{table}[t]
\centering
\scriptsize
\caption{In-domain benchmarking results (mIoU, \%) on the test sets under varying label ratios (1/16, 1/8, 1/4, and 1/2). Best values are bolded.}
\vspace{-0.1cm}
\begin{tabular}{l|cccc|cccc}
\toprule
\multirow{2}{*}{\textbf{Methods}} & \multicolumn{4}{c|}{\textbf{ResNet-18 + FCN}} & \multicolumn{4}{c}{\textbf{ViT-Tiny + FCN}} \\
& 1/16 & 1/8 & 1/4 & 1/2 & 1/16 & 1/8 & 1/4 & 1/2 \\
\midrule
\multicolumn{9}{c}{\textit{LUDB}} \\
\midrule
\textbf{Scratch}& 67.3& 71.3 & 72.9 & 74.1 & 66.0 & 71.3 & 75.9 & 78.5 
\\
\textbf{MT}& 70.8& 72.3& 73.6& 74.3& \textbf{73.6}& \textbf{76.7}& \textbf{78.7}& \textbf{80.2}\\
\textbf{FixMatch}& 70.9& 72.2& 72.9& 74.1& 72.2 & 76.6 & 78.1 & 80.1 
\\
\textbf{CPS}& 68.6& 71.6& 73.1& 74.3& 70.9 & 75.4 & 78.3 & 80.0 
\\
\textbf{ReCo}& \textbf{71.5}& \textbf{72.5}& 73.1 & 73.9 & 72.0 & 74.2 & 74.9 & 
75.0 \\
\textbf{ST++}& 69.2& 71.6& \textbf{73.7}& \textbf{74.5}& 71.3 & 76.0 & 78.0 & 80.1 \\
\midrule
\multicolumn{9}{c}{\textit{QTDB}} \\
\midrule
\textbf{Scratch}& 47.5 & 56.2 & 60.5 & 64.9 & 38.8 & 52.5 & 62.0 & 67.1 
\\
\textbf{MT}& 47.8 & 44.8 & 63.0 & 66.7 & \textbf{55.2}& \textbf{58.0}& 64.9 & 69.2 
\\
\textbf{FixMatch}& 46.7 & 53.3 & 58.2 & 66.3 & 46.7 & 53.4 & 64.2 & \textbf{69.3}\\
\textbf{CPS}& 53.2 & 57.1 & \textbf{64.7}& \textbf{68.0}& 48.9 & 55.3 & 58.9 & 67.4 
\\
\textbf{ReCo}& \textbf{53.4}& 53.4 & 58.7 & 64.5 & 45.8 & 47.6 & 59.4 & 62.9 
\\
\textbf{ST++}& 52.9 & \textbf{57.8}& 62.5 & 68.0 & 52.1 & 55.3 & \textbf{65.7}& 68.8 \\
\midrule
\multicolumn{9}{c}{\textit{ISP}} \\
\midrule
\textbf{Scratch}& 62.3 & 64.6 & 68.1 & 69.3 & 69.4 & 73.0& 76.3& 78.5 
\\
\textbf{MT}& 64.2 & 65.9 & 68.0 & 69.1 & \textbf{74.8}& \textbf{75.7}& 78.3 & 78.6 
\\
\textbf{FixMatch}& 63.1 & 64.4 & 67.9 & 68.9 & 74.7 & \textbf{75.7}& 78.0 & 79.4 
\\
\textbf{CPS}& \textbf{64.4}& \textbf{66.1}& \textbf{68.5}& 69.4 & 72.9 & 73.7 & \textbf{78.4}& \textbf{79.9}\\
\textbf{ReCo}& 62.3 & 64.7 & 67.4 & 68.1 & 71.3 & 73.2 & 76.3 & 77.2 
\\
\textbf{ST++}& 63.7 & 65.7 & \textbf{68.5}& \textbf{69.5}& 72.8 & 75.4 & 78.3 & 79.4 \\
\midrule
\multicolumn{9}{c}{\textit{Zhejiang}} \\
\midrule
\textbf{Scratch}& 76.7 & 78.9 & 80.7 & 82.3 & 69.5 & 73.3& 76.2& 80.2 
\\
\textbf{MT}& \textbf{79.2}& 80.1 & 81.5 & \textbf{82.9}& \textbf{79.1}& \textbf{81.4}& 81.6 & \textbf{83.6}\\
\textbf{FixMatch}& 79.0 & \textbf{80.4}& 81.2 & 82.7 & 76.1 & 80.2 & \textbf{82.0}& 83.5 
\\
\textbf{CPS}& 77.6 & 79.5 & 81.3 & 82.7 & 74.2 & 80.2 & 81.7 & 82.8 
\\
\textbf{ReCo}& 77.6 & 78.9 & 79.7 & 80.6 & 68.6 & 68.3 & 68.7 & 
63.3 \\
\textbf{ST++}& 78.8 & 79.8 & \textbf{81.9}& \textbf{82.9}& 73.9 & 79.0 & 81.3 & 83.5 \\
\bottomrule
\end{tabular}
\label{tab:bench-id}
\end{table}

\subsection{In-Domain Benchmarking Results}
The in-domain benchmarking performances are summarized in Table~\ref{tab:bench-id}. The results confirm the effectiveness of SemiSeg algorithms on ECG delineation.
The gap between SemiSeg algorithms and \textbf{Scratch} widened as the label proportion decreased (e.g., $1.7\%\text{p}\rightarrow2.8\%\text{p}\rightarrow5.4\%\text{p}\rightarrow7.6\%\text{p}$ in \textbf{MT} with ViT on \textit{LUDB}).
At the smallest label proportion (1/16), most of the algorithms demonstrated their effectiveness, indicating successful use of unlabeled data when labels were scarce.

From a database perspective, \textit{LUDB} had the fewest cases where SemiSeg algorithms underperformed \textbf{Scratch}, suggesting that their effectiveness was more consistently observable, likely due to the presence of accurate, lead-specific labels. From a model perspective, ViT-Tiny outperformed ResNet-18 overall. The detailed trends differ: ViT-Tiny consistently performed well with \textbf{MT}, while \textbf{ST++} remained competitive in ResNet-18. In general, \textbf{ReCo} showed limited benefit, notably on \textit{QTDB}, which exhibited performance degradation at all label ratios except for 1/16.

\begin{table}[th]
\centering
\scriptsize
\caption{
Cross-domain benchmarking results on the merged in-domain test set (\textit{LUDB}, \textit{QTDB}, \textit{ISP}, \textit{Zhejiang}). Models were trained on the merged in-domain labeled data and \textit{PTB-XL} unlabeled data. Best values are bolded ($\uparrow$: higher is better, $\downarrow$: lower is better).}
\vspace{-0.1cm}
\begin{tabular}{l|c|>{\centering\arraybackslash}p{1cm}|>{\centering\arraybackslash}p{0.8cm}>{\centering\arraybackslash}p{0.8cm}>{\centering\arraybackslash}p{0.8cm}}
\toprule
\multirow{2}{*}{\textbf{Methods}} & \multirow{2}{*}{\textbf{mIoU (\%) $\uparrow$}} &  \multicolumn{4}{c}{\textbf{MAE (ms) $\downarrow$}}\\
& &  Avg.&PR & QRS & QT \\
\midrule
\multicolumn{6}{c}{\textbf{ResNet-18 + FCN}} \\
\midrule
\textbf{Scratch}& \textbf{74.5}&  \textbf{20.6}&21.4 & 14.1 & 26.5 
\\
\textbf{MT}& 73.9 &  21.5 &22.7 & 15.6 & \textbf{26.3}\\
\textbf{FixMatch}& 73.6 &  21.5 &24.3 & \textbf{13.6}& 26.5 
\\
\textbf{CPS}& 74.4 &  21.1 &\textbf{21.2}& 15.0 & 27.2 
\\
\textbf{ReCo}& 73.7 &  22.3 &21.9 & 16.6 & 28.3 \\
\textbf{ST++}& 74.2 &  20.8 &21.5 & 14.2 & 26.8 \\
\midrule
\multicolumn{6}{c}{\textbf{ViT-Tiny + FCN}} \\
\midrule
\textbf{Scratch}& 81.7 &  18.5 &21.4 & 10.3 & 23.8 
\\
\textbf{MT}& 84.6 &  \textbf{14.9}&\textbf{13.1}& \textbf{9.7}& 21.9 
\\
\textbf{FixMatch}& 84.4 &  \textbf{14.9}
&13.4 & 9.9 & 21.4 
\\
\textbf{CPS}& 84.0 &  15.3 &15.4 & 10.2 & \textbf{20.4}\\
\textbf{ReCo}& 84.1 &  15.9 &13.8 & 10.4 & 23.5 
\\
\textbf{ST++}& \textbf{84.7}&  15.3 &14.4 & \textbf{9.7}& 21.9 \\
\bottomrule
\end{tabular}
\label{tab:bench-ood-internal}
\end{table}

\begin{table}[t]
\centering
\scriptsize
\caption{Out-domain generalization results on \textit{mECGDB}. Models were trained on the merged in-domain labeled data and \textit{PTB-XL} unlabeled data. Only MAE (ms) was reported, as \textit{mECGDB} contained interval labels (PR, QRS, QT) without delineation annotation. Best values are bolded.}
\vspace{-0.1cm}
\begin{tabular}{l|c|ccc|c|ccc}
\toprule
\multirow{2}{*}{\textbf{Methods}} &  \multicolumn{4}{c|}{\textbf{ResNet-18 + FCN}}&  \multicolumn{4}{c}{\textbf{ViT-Tiny + FCN}}\\
&  Avg.&PR & QRS & QT &  Avg.&PR & QRS & QT \\
\midrule
\textbf{Scratch}&  \textbf{26.4}&\textbf{26.0}& \textbf{18.7}& \textbf{34.6}&  23.2 &27.1 & 13.3 & 29.2 
\\
\textbf{MT}&  28.2 &27.1 & 20.6 & 36.9 
&  \textbf{20.7}&20.8 & \textbf{13.0}& 28.2 
\\
\textbf{FixMatch}        &  27.7 &28.1 & 19.5 & 35.4 
&  21.6 &21.4 & 13.8 & 29.5 
\\
\textbf{CPS}             &  28.1 &28.2 & 20.3 & 35.9 
&  21.1 &23.2 & 13.6 & \textbf{26.6}\\
\textbf{ReCo}            &  28.7 &26.1 & 22.8 & 37.3 &  21.0 &\textbf{19.5}& 13.2 & 30.3 
\\
\textbf{ST++}            &  27.8 &27.4 & 19.5 & 36.6 &  21.6 &23.4 & 13.6 & 27.8 \\
\bottomrule
\end{tabular}
\label{tab:bench-ood-external}
\end{table}

\subsection{Cross-Domain Benchmarking Results}
The benchmarking performance patterns in the cross-domain setting deviated from those in the in-domain setting. SemiSeg algorithms provided limited benefit when using ResNet-18 in the cross-domain setting as shown in Table~\ref{tab:bench-ood-internal} and Table~\ref{tab:bench-ood-external}. All the SemiSeg algorithms failed to outperform the \textbf{Scratch}, indicating poor generalization. In contrast, ViT-Tiny consistently benefited from SemiSeg algorithms. \textbf{MT}, \textbf{FixMatch}, and \textbf{ST++} achieved remarkable gains, with \textbf{ST++} yielding the highest mIoU (84.7\%) and \textbf{MT}/\textbf{FixMatch} showing the lowest average ECG interval MAE (14.9 ms).

Notably, FixMatch with a ViT backbone, one of the best models on the merged in-domain test set, did not lead on \textit{mECGDB}, surpassed by \textbf{MT} in the average MAE ($20.7\text{ ms}<21.6\text{ ms}$). It likely reflected the distribution shift between clinical 12-lead ECGs and mobile ECGs collected outside clinical settings. The models may have overfitted to the large-scale unlabeled set (\textit{PTB-XL}), limiting cross-domain robustness. Furthermore, the model with the highest mIoU did not produce the lowest average MAE. These findings highlight the need for domain adaptation techniques and a multi-metric evaluation that considers both segmentation quality and clinical interval accuracy.

\section{Conclusion}
We presented \textbf{\textsf{SemiSegECG}}, a unified benchmark and evaluation protocol for semi-supervised ECG delineation. Across diverse public datasets, SemiSeg algorithms consistently improved delineation performances under label scarcity, with ViT backbones providing the clearest gains. Meanwhile, the inconsistency in model performance across datasets between the in-domain merged test set and \textit{mECGDB} highlights the impact of distribution shifts and the need for domain-aware training and ECG-specific augmentation strategies. Limitations include the modest size of annotated datasets and the scope of evaluated methods, which focused on representative SemiSeg algorithms originally developed for computer vision. Future work may leverage the \textbf{\textsf{SemiSegECG}} to explore more advanced or ECG-specific semi-supervised learning approaches and domain adaptation techniques tailored to the challenges of physiological signals.

{
    \small
    \bibliographystyle{ieeenat_fullname}

}

\end{document}